\documentclass[runningheads]{llncs}

\usepackage{eccv}

\usepackage{eccvabbrv}

\usepackage{graphicx}
\usepackage{booktabs}

\usepackage[accsupp]{axessibility}  

\usepackage{hyperref}

\usepackage{orcidlink}

\usepackage{multirow}
\usepackage{multicol}
\usepackage{booktabs}
\usepackage{amsmath, amssymb}
\usepackage{type1cm}
\usepackage{comment}
\usepackage{cuted}
\usepackage{adjustbox}

\usepackage{tabularx}

\usepackage[table]{xcolor}
\definecolor{mypink}{HTML}{ff6b6b}
\definecolor{mygray}{HTML}{e9ecef}

\def\etal{\textit{et al. }}

\def\model{SoundMHPE }
\def\modelnospace{SoundMHPE}
\def\modelall{(Sound-based Multi-person Human Pose Estimator)}

\begin{document}

\title{Sound-based Multi-Person 3D Pose Estimation} 

\titlerunning{Sound-based Multi-Person 3D Pose Estimation}

\author{
Yusuke Oumi\inst{1}\orcidlink{0009-0009-8123-9796}
\and
Yuto Shibata\inst{1}\orcidlink{0009-0005-4225-3887}
\and
Go Irie\inst{1,2}\orcidlink{0000-0002-4309-4700}
\and \\
Akisato Kimura\inst{3}\orcidlink{0009-0007-3042-6810}
\and
Yoshimitsu Aoki\inst{1}\orcidlink{0000-0001-7361-0027}
\and
Mariko Isogawa\inst{1}\orcidlink{0000-0001-9560-0276}
}

\authorrunning{Y.~Oumi et al.}


\institute{
Keio University, Yokohama, Kanagawa 223-8522, Japan \and
Tokyo University of Science, Katsushika, Tokyo 125-8585, Japan\and
NTT, Inc., Keihanna Science City, Kyoto 619-0237, Japan 
}


\maketitle

\begin{abstract}

Can we recover the 3D poses of multiple people using only sound?
This paper presents the first attempt to estimate multi-person 3D poses solely from acoustic signals.
Estimating the poses of multiple individuals using acoustic signals is inherently challenging due to the superposition of motion-dependent signal variations. Unlike single-person scenarios, the presence of multiple subjects leads to overlapping acoustic signatures, making it difficult to attribute specific signal changes to an individual's pose. Furthermore, the complexity is compounded by inter-person reflections, which introduce intricate propagation delays that obscure the temporal motion-acoustic relationship.
To address these issues, we propose \model ~\modelall, a novel encoder-decoder framework consisting of two key components. First, the Acoustic Multi-scale Encoder captures diverse temporal and fine-grained frequency features to isolate subtle acoustic signatures from complex, overlapping signals. Second, the Temporal Pose Decoder employs an attention mechanism to disentangle multi-person information across successive frames. 
By jointly accounting for temporal dynamics and inter-person dependencies, this component precisely reconstructs frame-wise individual poses.
To validate our approach, we constructed the 6-hour \emph{Acoustic Multi-person Pose (AMP)} dataset consisting of 432K synchronized frames of multi-person pose and acoustic data, and demonstrated that our \model outperforms baseline models.
Project page: https://oumi03.github.io/sound-mhpe/

\end{abstract}

\begin{figure}[t]
  \centering
    \includegraphics[width=0.94\linewidth]
    {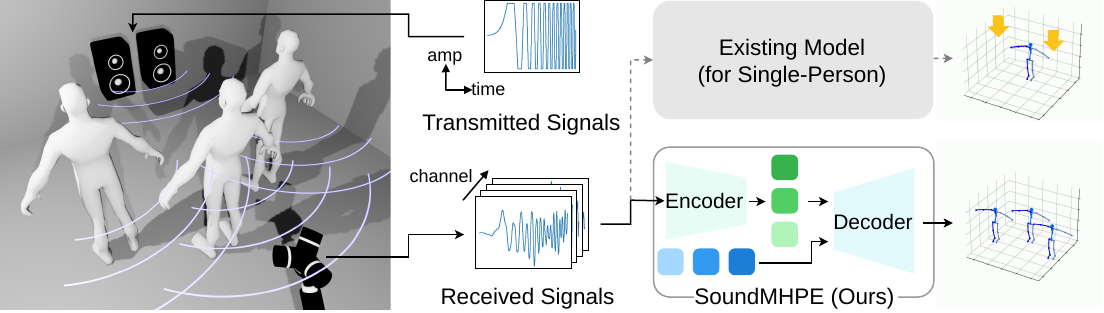}

   \caption{We propose \modelnospace, a sound-based multi-person 3D pose estimation method. 
    Our system adopts an active acoustic sensing approach, where a speaker emits a transmitted signal and the received signal is used for the model input. 
    While existing acoustic pose estimation models are limited to single-person estimation, \model enables simultaneous estimation of multiple individuals.}
   \label{fig:teaser} 
\end{figure}

\section{Introduction}
\label{sec:intro}

Understanding human poses in complex, real-world environments is a long-standing fundamental problem in computer vision~\cite{survey-2Dmulti_wang_2021,survey-monocular-multi_dos_2021}. In everyday scenarios, such as crowded city streets, workplaces, or indoor facilities, multiple people naturally coexist and interact.
Therefore, obtaining multi-person pose information in a non-invasive manner is beneficial for a wide range of applications, including monitoring systems~\cite{humanfall_raza_2025}, sports analysis~\cite{systematic_badiola_2021}, and disaster relief efforts~\cite{throughwall_song_2021}.

Multi-person pose estimation has primarily been studied using RGB images~\cite{openpose_tpami,rmpe_fang_2017, PETR_shi_2022},  
but these approaches are highly susceptible to occlusion and low-light conditions~\cite{low-light_lee_2023}.
In contrast, 
wireless-signal-based methods~\cite{RF-pose,PETR-wifi_yan_2024,raptr_kato_2025,IR-UWB_kim_2023} can be used in dark environments and can estimate the poses of occluded individuals by passing through obstacles.
However, wireless signals are still susceptible to occlusion caused by water or metal~\cite{RF-pose}.
To address these limitations, active acoustic sensing based pose estimation has recently gained attention~\cite{listening_shibata_2023,sound2pose-pos_Oumi_2024,generalizing_nakamura_2024,bgm2pose_shibata_2025, echomotion_mosuily_2025}.
Acoustic signals are unaffected by lighting conditions and, due to their relatively long wavelengths, offer the potential for estimating poses even behind obstacles such as metal.
While these approaches can estimate the poses using only acoustic signals, they are limited to estimating poses for a single person.


To fill this gap, we take the first step toward estimating the poses of multiple individuals based solely on acoustic signals (\cref{fig:teaser}). 
In contrast to single-person settings, where acoustic variations are uniquely determined by an individual's movements, the multi-person scenario introduces substantial signal ambiguity. Specifically, the observed signals represent a superposition of concurrent motion features, which is further complicated by non-trivial acoustic propagation delays caused by multi-body reflections. Together, these phenomena obscure the direct correspondence between the subjects' physical poses and the acoustic data.

To address these challenges, we propose \model~\modelall,  a novel encoder-decoder framework that explicitly accounts for the complexity of acoustic signals caused by multiple subjects.
To disentangle dynamic motion-dependent information from acoustically overlapped features and complex multi-body reflections, we propose an Acoustic Multi-scale Encoder.
This approach employs a multi-resolution STFT with varying window sizes and effectively leverages the complementary strengths of different temporal and frequency scales. Short-window spectrograms capture high-fidelity temporal dynamics, while long-window counterparts resolve fine-grained frequency variations. This dual-representation ensures a comprehensive characterization of the signal across both the time and frequency domains.
In addition, to explicitly model both inter-person interactions and intra-person temporal dynamics, we introduce the Temporal Pose Decoder. This module employs a spatio-temporal query mechanism where dedicated queries are assigned to each individual across successive frames. Such a formulation seamlessly integrates the temporal evolution of individual poses with the global context of multi-body interactions.

Furthermore, since multi-person pose estimation using acoustic signals has not been previously explored, we construct our 6-hour \emph{AMP dataset} (Acoustic Multi-person Pose), consisting of synchronized multi-person poses and acoustic signals.
Extensive experiments on the \emph{AMP dataset} show that \model achieves superior performance over baseline methods, and detailed ablation studies further confirm the effectiveness of our proposed components.

In summary, the contributions of our paper are as follows.
(1) We tackle the first approach for multi-person pose estimation using acoustic signals.
(2) We introduce an Acoustic Multi-scale Encoder to capture both fine-grained temporal dynamics and subtle acoustic variations in the acoustic signals.
(3) We introduce the Temporal Pose Decoder to jointly model intra-person temporal dynamics and inter-person acoustic interactions.
(4) Since this task has not been previously explored, we constructed our AMP dataset and conducted experiments to evaluate our proposed \model model.

\begin{table*}[t]
    \centering
        \caption{Comparison of existing human pose estimation methods and our method}
        \begin{center}\resizebox{1.0\linewidth}{!}{ 
        \begin{tabular}[t]{rcccc}
        \toprule
        Method & Modality & Restrictions & Estimation target 
        \\
        \midrule
        RGB-based~\cite{openpose_tpami,rmpe_fang_2017, PETR_shi_2022} & RGB images/videos & \color{Red}{Dark environment, occlusion} & \color{ForestGreen}{Single/Multiple persons}\\
        Wireless-signal-based~\cite{PETR-wifi_yan_2024,raptr_kato_2025, IR-UWB_kim_2023} & RF/WiFi, mmWave, UWB& \color{Red}{Occlusion, precision equipment} & \color{ForestGreen}{Single/Multiple persons}\\
        Acoustic-signal-based~\cite{listening_shibata_2023,sound2pose-pos_Oumi_2024} & Audio& Soundproof room & \color{Red}{Only single person}
&\\
        \rowcolor{mygray}
        Ours & Audio&  Soundproof room & \color{ForestGreen}{Single/Multiple persons}\\
        \bottomrule
        \label{tab:related_work}
        \end{tabular}
        }    
        \end{center}
        
\end{table*}

\section{Related Work}
\label{sec:related}

\subsection{Non-invasive Multi-person Pose Estimation}
\label{sec:21}
Multi-person pose estimation is a traditional task in computer vision, and
many effective methods using
RGB images~\cite{openpose_tpami, PETR_shi_2022, grouppose_liu2023} and videos~\cite{psvt_qiu_2023,snipper_zou_2023},
radio frequency (RF)/WiFi signals~\cite{person-in-wifi_wang_2019,PETR-wifi_yan_2024,RF-pose},
mmWave~\cite{m3track,multi-mmwave_feng_2025, raptr_kato_2025},
and UWB radar signals~\cite{IR-UWB_kim_2023,radarformer_zheng_2023, joint-loss_huang_2025}
have been proposed.
However, both RGB images and wireless signals (RF/WiFi, mmWave, UWB) 
each have 
scenes 
in which estimation using them becomes less effective (see Table~\ref{tab:related_work}).
RGB-based methods suffer from occlusion issues and decreased estimation accuracy in low-light conditions~\cite{low-light_lee_2023}. 
Additionally, 
RGB-based approaches are prone to raising privacy concerns~\cite{privacy-preserving_hinojosa_2021}.
Wireless-signal-based methods can estimate occluded targets by penetrating certain obstacles, but they
face challenges in dealing with obstruction by water or metal~\cite{RF-pose}.
Furthermore, the methods using wireless signals are limited in environments where wireless communication is restricted, such as medical facilities or aircraft.
We address these challenges by 
using acoustic signals.

\subsection{Active Acoustic Sensing for Human Pose Estimation}
\label{sec:22}

Active acoustic sensing estimates target states by emitting sound signals and analyzing the received acoustic signals.
For non-invasive human state estimation, this technique has been applied to tasks such as action recognition~\cite{hear_tanigawa_2024} and mesh reconstruction combined with RGB images~\cite{sonicmesh_liang_2024}.
More recently, active acoustic sensing has also been extended to human pose estimation. In particular, prior works have explored pose estimation with chirp signals~\cite{listening_shibata_2023,sound2pose-pos_Oumi_2024}, inaudible continuous tones~\cite{echomotion_mosuily_2025}, and even music~\cite{generalizing_nakamura_2024,bgm2pose_shibata_2025}.
However, all these
methods are limited to single-person settings and cannot handle multiple 
persons
(Table \ref{tab:related_work}).

\subsection{Spatio-Temporal Modeling for Multi-Person Pose Estimation}
Spatio-temporal modeling is important for multi-person pose estimation, as temporal dynamics and inter-person dependencies must be jointly captured. For time-series data such as acoustic signals and videos, models that leverage temporal information are commonly employed. 

In acoustic signal-based human pose estimation, Shibata~\etal~\cite{listening_shibata_2023} proposed a CNN-based architecture with temporal convolutions, while Oumi~\etal~\cite{sound2pose-pos_Oumi_2024} enhanced temporal modeling by incorporating prior information about acoustic signals. However, these methods deal with single-person settings and primarily focus on temporal modeling. 

In video-based multi-person pose estimation, a range of modeling strategies for spatio-temporal representations has been explored, including temporal-CNN~\cite{temp-conv_pavllo_2019}, LSTM~\cite{lstm-pose_luo_2018}, and Transformer~\cite{pose-former_zheng_2021}. More recently, DETR-like architectures~\cite{DETR_carion_2020} have been extended to multi-person pose estimation~\cite{psvt_qiu_2023,snipper_zou_2023}, where object queries enable explicit modeling of individual instances. Notably, Snipper~\cite{snipper_zou_2023} employs frame-wise and future-frame queries to jointly perform pose estimation and forecasting for each person. Inspired by these spatio-temporal query-based designs, we construct a sequence of queries for each individual, enabling the model to jointly learn the relationship between multi-frame acoustic features and the corresponding multi-frame poses across individuals.

\begin{figure}[t]
  \centering
   \includegraphics[width=\linewidth]{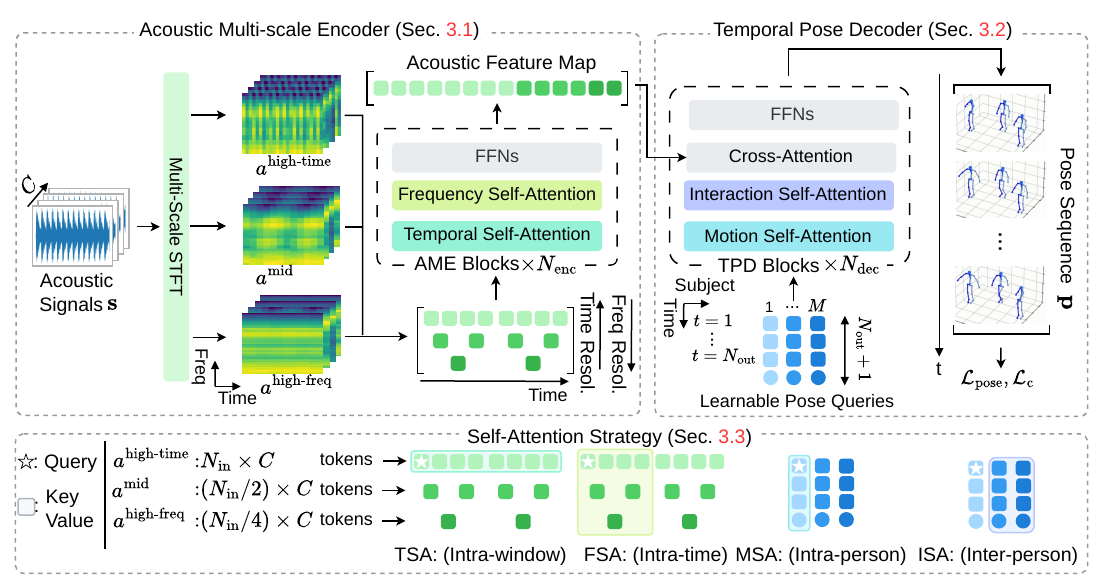}
   \caption{Proposed framework for sound-based multi-person pose estimation.
    (Top left) \model first employs an Acoustic Multi-scale Encoder to generate spectrograms with diverse time--frequency characteristics and obtain an acoustic feature map. 
    (Top right) Subsequently, in the Temporal Pose Decoder, we assign each individual a set of learnable pose queries. (Bottom) These encoder and decoder modules leverage customized self-attention to jointly model spatio-temporal dynamics, multi-resolution features, and inter/intra-person pose relationships.
   }
   \label{fig:proposed_framework} 
\end{figure}

\section{Methodology}
\label{sec:method}

As shown in~\cref{fig:proposed_framework}, 
\model estimates the 3D pose sequence of multiple subjects 
$\mathbf{p_j} = \{p_{j,t}\}_{t=1}^T$
from an acoustic signal
$\mathbf{s} = \{s_t\}_{t=1}^{T \times L}$.
Here, $j$ represents subject ID, $T$ denotes the sequence length of 
$\mathbf{p_j}$, and $L$ is the length of the acoustic signal sequence corresponding to single frame pose $p_{j,t}$. 
The following subsections discuss the key components of \modelnospace.
We introduce our two main technical contributions, Acoustic Multi-scale Encoder~(\cref{sec:31}) and Temporal Pose Decoder (\cref{sec:32}). 
In~\cref{sec:33}, we present detailed implementation of the self-attention mechanism used in \modelnospace.

\subsection{Acoustic Multi-scale Encoder}
\label{sec:31}
In our framework, the input acoustic signal $\mathbf{s}$ is captured using active acoustic sensing. 
We first apply multi-scale Short Time Fourier Transform (STFT) to obtain a multi-scale log-Mel spectrogram. This spectrogram is tokenized and fed into an Acoustic Multi-scale Encoder (AME), producing an acoustic feature map.
\model simultaneously estimates $N_\mathrm{out}$ consecutive pose frames. Following~\cite{sound2pose-pos_Oumi_2024}, to appropriately capture the temporal relationships in the acoustic signals, we use the acoustic signals corresponding to $N_{\mathrm{out}} + N_\mathrm{prev}$ pose frames to predict the latter $N_{\mathrm{out}}$ frames. 
Therefore, \model estimates the pose sequence $\{p_{j,t}\}_{t=1}^{N_{\mathrm{out}}}$ from the acoustic signal $\{s_t\}_{t=1-N_\mathrm{prev}\times L}^{N_\mathrm{out}\times L}$. We define $N_{\mathrm{in}} = N_{\mathrm{out}} + N_\mathrm{prev}$.

\noindent
\textbf{Active Acoustic Sensing.}
The acoustic signal $\mathbf{s}$ is captured using a pair of speakers and a microphone. To capture the three-dimensional spatial information, we use an ambisonics microphone that records across four channels, i.e., $\mathrm{W}$ as an omnidirectional channel, and $\mathrm{X, Y, Z}$, which capture directional components. Following existing acoustic-sensing-based human pose estimation methods~\cite{listening_shibata_2023,sound2pose-pos_Oumi_2024}, we use a time stretched pulse (TSP) signal, which is a periodic signal whose frequency changes within each cycle, as the emitted sound source.

\begin{figure}[t]
  \centering
   \includegraphics[width=0.9\linewidth]{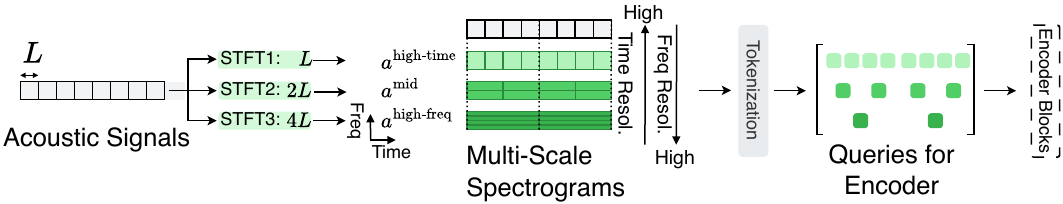}
    \caption{
    By performing STFT with three different temporal window sizes, $L$, $2L$, and $4L$, we generate spectrograms denoted as $a^{\mathrm{high\text{-}time}}$, $a^{\mathrm{mid}}$, and $a^{\mathrm{high\text{-}freq}}$, corresponding to high-temporal, intermediate, and high-frequency resolutions, respectively.
   }
   \label{fig:stft}
\end{figure}

\noindent
\textbf{Log-Mel Spectrogram.}
We convert the acoustic signal $\mathbf{s}$ into a log-Mel spectrogram.
The received signal is segmented channel-wise with a fixed interval $L$, and each segment is transformed into a spectrogram via STFT.
Then, the spectrogram is projected onto the Mel scale and converted to log scale 
as follows:
\begin{eqnarray}
    a_{t,c} = \log (\,H_\mathrm{mel} \cdot \mathcal{F}(\{s_{t',c}\}^{t\times L}_{t'=(t-1)\times L+1})\,)\text{,}
\end{eqnarray}
where $H_\mathrm{mel}$ denotes the Mel filter banks, $\mathcal{F}$ represents the Fourier transform operation, and $c$ indicates the channel index.
From an acoustic signal of length $T \times L$, we obtain log-Mel spectrogram $\mathbf{a} = \{a_{t}\}_{t=1}^{T}$, and each element $a_{t} \in \mathbb{R}^{(C,B)}$, 
where $C$ denotes the number of microphone channels (four in this paper), and $B$ denotes the number of Mel filter banks.

\noindent
\textbf{Multi-Scale STFT.}
Following~\cite{stfnets_yao_2019}, to
leverage high resolution in both the temporal and frequency domains, we extract the log-Mel spectrogram using various window sizes: $L$, $2L$, and $4L$ paired with Mel filter banks $B$, $2B$, and $4B$, respectively.
Given an acoustic signal of length $N_{\mathrm{in}} \times L$, these configurations produce multi-scale spectrograms: $a^\mathrm{high\text{-}time} \in \mathbb{R}^{(N_{\mathrm{in}}, C, B)}$, $\allowbreak a^\mathrm{mid} \in \mathbb{R}^{(N_{\mathrm{in}}/2, C, 2B)}$, and $a^\mathrm{high\text{-}freq} \in \mathbb{R}^{(N_{\mathrm{in}}/4, C, 4B)}$ (see~\cref{fig:stft}).
$a^\mathrm{high\text{-}time}$ denotes a spectrogram with high temporal resolution, while $a^\mathrm{high\text{-}freq}$ denotes a spectrogram with high frequency resolution. $a^\mathrm{mid}$ corresponds to an intermediate resolution between the two.

\noindent
\textbf{Implementation of the Encoder.}
The multi-scale spectrograms are concatenated along the temporal axis, after which linear layers are applied independently at each time step to unify their frequency dimensions. This 
yields a tensor with the size of
$((7/4)\times N_{\mathrm{in}}, C, E)$,
where $E$ denotes the embedding dimension of the model.
The tensor is subsequently flattened along the time--channel dimensions to form the input queries of the encoder.
Our AME consists of 
$N_\mathrm{enc}$ stacked AME blocks, each composed of a Temporal Self-Attention, a Frequency Self-Attention and feed-forward networks (FFNs), and outputs an acoustic feature map.
The details of the two attention modules are described in~\cref{sec:33}.

\subsection{Temporal Pose Decoder}
\label{sec:32}
This section introduces the Temporal Pose Decoder (TPD), which explicitly learns the temporal relationship between acoustic features and poses of each frame.
We first present the preliminary concepts necessary to understand our method, followed by an explanation of the query design
and loss functions.

\noindent
\textbf{Preliminary.}
DETR is a Transformer-based model designed for end-to-end object detection~\cite{DETR_carion_2020}.
In their framework, the decoder prepares $M$ queries, each corresponding to a potential object.
Then, by computing cross-attention between these queries and the feature maps
produced by the encoder, the decoder 
extracts features corresponding to each object individually from the feature map.
In DETR, the loss is computed by matching predicted objects with ground truth objects using the Hungarian matching algorithm.
A similar approach is used in DETR-based multi-person pose estimation models~\cite{PETR_shi_2022,PETR-wifi_yan_2024},
where the 
decoder predicts a single-frame pose $\hat{p_i}$ and a confidence score ${\hat{c_i}}$ for each query.
During inference, only the poses whose confidence ${\hat{c_i}}$ exceeds a predefined threshold are output as the final results.

\begin{figure}[t]
  \centering
   \includegraphics[width=0.8\linewidth]{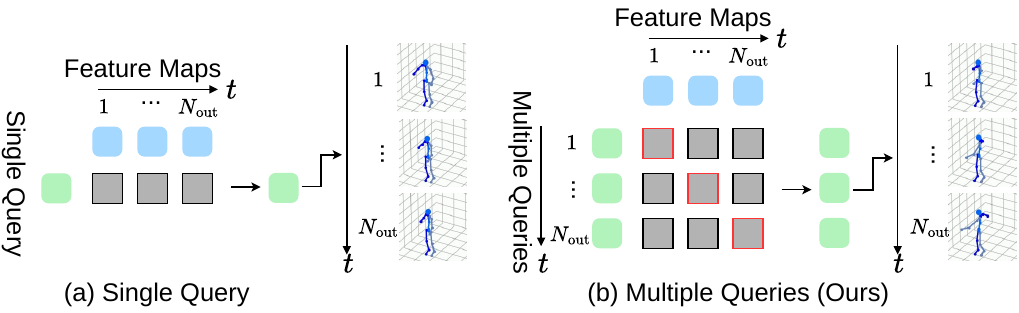}
    \caption{
   Cross-attention between queries and acoustic feature maps in TPD. (a) Single Query for Multi-frame poses: The pose information from multiple frames is aggregated into a single query. (b) Multiple Queries for Multi-frame poses: Multiple queries enable the model to focus on the acoustic features of different frames for poses in each frame.
   }
   \label{fig:cross_attn}
\end{figure}

\noindent
\textbf{Multiple-Queries for a Single Person. }
\model is designed based on the aforementioned DETR-based architecture, which estimates the poses of multiple individuals by extracting information related to each person from the acoustic feature maps through cross-attention.
Additionally, following existing sound-based pose estimation models~\cite{listening_shibata_2023}, we estimate multi-frame poses 
simultaneously 
rather than a single-frame pose to ensure smooth transitions between adjacent frames.
Therefore, in 
a standard DETR-style approach where $M$ queries represent $M$ potential individuals, each query must compress the entire temporal sequence of a person’s pose into a single representation. As shown in \cref{fig:cross_attn} (a), this collapses the temporal dimension, making it difficult to capture fine-grained acoustic details such as inter-person reflections and their precise timing.
To address this issue, we propose Temporal Pose Decoder (TPD), which uses $N_\mathrm{out}$ pose queries 
for each individual to estimate the multi-frame poses. By preparing $N_\mathrm{out}$ queries, 
each assigned to an estimated pose frame,
the queries corresponding to earlier poses can attend to earlier acoustic features, while those corresponding to later poses can focus on the acoustic features of subsequent time steps (\cref{fig:cross_attn} (b)).
To effectively disentangle individual motions across time, each query is conditioned on both a temporal positional embedding and a subject-specific embedding.
In addition, since the confidence score $\hat{c_i}$ for each of the 
$M$ instances is independent of a specific frame, we follow the approach of~\cite{grouppose_liu2023} and introduce $M$ additional instance queries.
The confidence score $\hat{c_i}$ is obtained by passing the decoder output associated with the instance query through a linear layer.
Therefore, the number of learnable queries in TPD becomes $M \times (N_\mathrm{out} + 1)$.
Our TPD architecture is composed of $N_\mathrm{dec}$ stacked TPD blocks, each composed of Motion Self-Attention (MSA), Interaction Self-Attention (ISA), Cross-Attention and FFNs.
The details of these self-attention modules are described in~\cref{sec:33}.

\noindent
\textbf{Loss Functions.}
We utilize the Hungarian algorithm for matching-based loss calculation like \cite{DETR_carion_2020}.
The total loss $ \mathcal{L}$ consists of two loss functions:
$\mathcal{L}_\mathrm{pose}$ and $\mathcal{L}_\mathrm{c}$.
\begin{eqnarray}
    \mathcal{L} = \mathcal{L}_\mathrm{pose}+\lambda\mathcal{L}_\mathrm{c},
\end{eqnarray}
where $\mathcal{L}_\mathrm{pose}$ is the mean squared error loss between the ground truth pose and the predicted pose matched to  the GT pose. 
$\mathcal{L}_\mathrm{c}$ is the binary cross entropy loss and is used to learn the confidence score.
$\lambda$ is a weight hyperparameter.

\begin{figure}[t]
  \centering
   \includegraphics[width=0.9\linewidth]{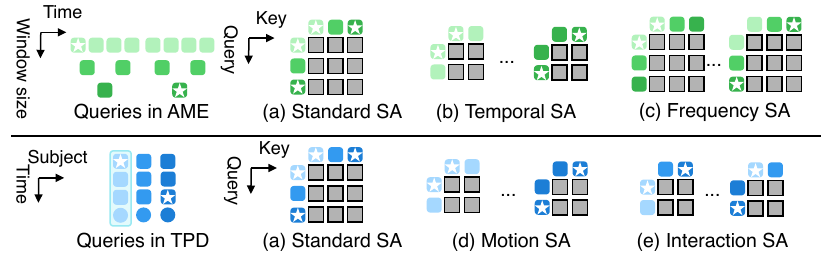}
    \caption{
    Self-Attention in AME and TPD.
    (a) Standard Self-Attention: Compute attention across all queries. 
    (b) Temporal Self-Attention: Models temporal dependencies within spectrograms generated using the same STFT window size.
    (c) Frequency Self-Attention: Captures relationships among spectrograms derived from the same acoustic sequence with different window sizes.
    (d) Motion Self-Attention: Learns temporal relationships of the same individual across frames.
    (e) Interaction Self-Attention: Models relationships between different individuals to capture inter-person interactions.
   }
   \label{fig:self_attn}
\end{figure}

\subsection{Self-Attention Strategy}
\label{sec:33}
The self-attention mechanisms in \model are specifically designed to disentangle the complex spatio-temporal features inherent in multi-scale spectrograms. As demonstrated in our ablation studies (Sec.~\ref{sec:52}), decoupling temporal and frequency dependencies is essential for maintaining estimation precision in multi-person environments.
In the following, we describe the detailed implementations of the self-attention used in our Acoustic Multi-scale Encoder (AME) and Temporal Pose Decoder (TPD), respectively.

\noindent
\textbf{Self-Attention in AME. }
Our AME performs Multi-Scale STFT using three temporal window sizes and we extracts $a^\mathrm{high\text{-}time} \in \mathbb{R}^{(N_{\mathrm{in}}, C, B)}$, $\allowbreak a^\mathrm{mid} \in \mathbb{R}^{(N_{\mathrm{in}}/2, C, 2B)}$, and $a^\mathrm{high\text{-}freq} \in \mathbb{R}^{(N_{\mathrm{in}}/4, C, 4B)}$.
The queries generated from these spectrograms contain information from different time and frequency resolutions. 
As shown in \cref{fig:self_attn} (a), standard self-attention merges queries from all time steps and resolutions into a single operation. This simultaneous processing hinders the model's ability to decouple temporal relationships from multi-scale frequency information, leading to less effective feature extraction.
To explicitly model these characteristics, we introduce two types of self-attention mechanisms: Temporal Self-Attention (TSA) and Frequency Self-Attention (FSA).
TSA performs self-attention within the features generated from each specific STFT window resolution (\cref{fig:self_attn} (b)), focusing on the temporal relationships within each spectrogram.
Specifically, self-attention is calculated over $N_{\mathrm{in}} \times C$, $(N_{\mathrm{in}}/2)\times C$, and $(N_{\mathrm{in}}/4) \times C$ queries corresponding to $a^{\mathrm{high\text{-}time}}$, $a^{\mathrm{mid}}$, and $a^{\mathrm{high\text{-}freq}}$, respectively.
FSA computes attention across spectrograms generated with different temporal window sizes, but originating from the same acoustic signal sequence (\cref{fig:self_attn} (c)). 
Specifically, since temporal windows of $L$, $2L$, and $4L$ are employed, four queries from $a^{\mathrm{high\text{-}time}}$, two queries from $a^{\mathrm{mid}}$, and one query from $a^{\mathrm{high\text{-}freq}}$ correspond to the same acoustic sequence of length $4L$. 
Therefore, self-attention is computed over these $7\times C$ queries.

\begin{figure}[t]
  \centering
   \includegraphics[width=0.9\linewidth]{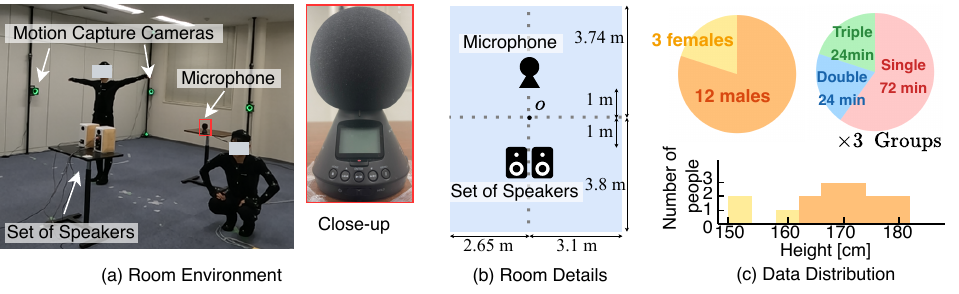}
   \caption{Experimental setup. (a,b) Our measurement environment consists of a set of speakers and a microphone for active acoustic sensing, along with motion capture cameras to obtain ground-truth poses.
   (c) Our AMP dataset consists of 12 male and 3 female participants, with heights ranging from 150\,cm to 181\,cm. The participants were divided into three groups. For each group, we collected 72 minutes of single-person data, 24 minutes of double-person data, and 24 minutes of triple-person data.}
   \label{fig:setup}
\end{figure}

\noindent
\textbf{Self-Attention in TPD.}
To efficiently extract pose features in multi-person scenarios, our TPD decouples the self-attention process into two distinct axes: (i) intra-person temporal dynamics and (ii) inter-person interactions.
While a standard DETR-based decoder employs a single query per person to model inter-instance relationships, our TPD assigns $N_\mathrm{out}+1$ queries to each individual to capture fine-grained temporal changes. Applying global self-attention across all $M \times (N_\mathrm{out}+1)$ queries would indiscriminately mix temporal cues with inter-person reflective dependencies, hindering the model's ability to focus on either.
To resolve this, we introduce the Motion Self-Attention (MSA) and Interaction Self-Attention (ISA). The MSA restricts its attention to the $N_\mathrm{out}+1$ queries belonging to the same individual. 
By focusing solely on an individual's own queries, the model effectively extracts consistent intra-person temporal dynamics (\cref{fig:self_attn} (d)).
In contrast, the ISA computes attention between the $i$-th subject and the remaining $(M-1) \times (N_\mathrm{out}+1)$ queries of other subjects. This allows the model to explicitly capture inter-person dependencies and mutual acoustic influences (\cref{fig:self_attn} (e)).

\section{Experimental Settings}
\label{sec:setting}
\subsection{Multi-Person Acoustic Pose (AMP) Dataset}
\label{sec:41}

Since we are tackling a novel task for which no existing dataset is available, we constructed the 6-hour Acoustic Multi-person Pose (AMP) dataset.
This dataset consists of synchronized acoustic data acquired through active acoustic sensing and 3D coordinate data of multiple individuals.

{\bf{The measurement environment}} is shown in~\cref{fig:setup} (a,b).
We employed a pair of loudspeakers (Edifier ED-S880DB) and an ambisonics microphone (Zoom H3-VR) for acoustic data collection in an indoor room where background noise and reverberation were present. A Motive motion capture system (OptiTrack) equipped with 16 cameras was utilized for obtaining ground-truth pose data.

{\bf{The AMP dataset statistics}} are shown in~\cref{fig:setup} (c).
It consists of 15 subjects (12 male, 3 female) with heights ranging from 150 to 181 cm.
To prepare a multi-person dataset, these subjects were divided into three groups. Within each group, subject pairings and positions were randomized during data collection. 
Participants performed a variety of poses, including walking, twisting, and raising both hands, in a random order and  at random speeds.
We used a skeleton consisting of 21 joints. Each joint is represented by the head, neck, shoulders, arms, forearms, hands, waist, thighs, shins, feets, toes, hip, and spine.
For each group, we collected 72 minutes of single-person data, and 24 minutes each of double-person and triple-person data.
Ground-truth poses were recorded at 20 fps, resulting in a total of approximately 432K frames.

\subsection{Baseline Methods}
\label{sec:42}
Since no prior work exists for acoustic multi-person pose estimation, we adapted two closely related models as baselines:
(1) Adapted Shibata \textit{et al.}~\cite{listening_shibata_2023}: Although originally for single-person estimation, this work achieved precise active pose estimation with the TSP signal. To enable a fair comparison, we extended its feature extractor with a multi-person regression head.
(2) Repurposed Yan \textit{et al.}~\cite{PETR-wifi_yan_2024}: We adapted this multi-person WiFi-based model because WiFi CSI and acoustic log-Mel spectrograms are functionally analogous; both capture environment-induced signal perturbations in the time--frequency domain. We modified the input stem to accommodate an acoustic spectrogram while preserving the original transformer architecture.
Both baselines were retrained from scratch on our dataset to ensure fair comparison.

\subsection{Evaluation Metrics}
\label{sec:43}
We employ three evaluation metrics: mean per joint position error (MPJPE), Procrustes-aligned mean per joint position error (PA-MPJPE), and percentage of correct keypoints (PCK).
MPJPE is computed as the mean Euclidean distance between the predicted and ground-truth joint positions.
PA-MPJPE first aligns the predicted pose to the ground-truth pose using Procrustes analysis, which removes differences in global translation, rotation, and scale. MPJPE is then computed on the aligned poses. 
PCK measures the percentage of joints whose euclidean distance to the ground truth is within a predefined threshold.
We used PCKh@0.5, where the threshold is half the distance between the head and neck.

\begin{figure}[t]
\begin{center}    
\begin{minipage}[t]{0.45\linewidth}
    \centering
        \captionof{table}{Comparison against baselines}
        \begin{adjustbox}{width=\linewidth}
        \begin{tabular}[t]{@{\hskip 1mm}lccc@{\hskip 1mm}cccccccccccc}
        \toprule
        \multirow{3}{*}{Method} 
         & \multirow{2}{*}{\shortstack[c]{MPJPE \\ \lbrack mm\rbrack}} & \multirow{2}{*}{\shortstack[c]{PA-MPJPE \\ \lbrack mm\rbrack}} & \multirow{2}{*}{\shortstack[c]{PCKh \\ @0.5}}\\
        \\
        & ($\downarrow$) & ($\downarrow$) & ($\uparrow$)
        \\
        \midrule
        Shibata \textit{et al.} ~\cite{listening_shibata_2023} & 121.7 & 71.5 & 0.36\\
        Yan \textit{et al.} ~\cite{PETR-wifi_yan_2024} & 119.9 &  69.7 &0.36\\
        \rowcolor{mypink!20}
        Ours & \textbf{106.5} & \textbf{65.0} & \textbf{0.43} \\
        \bottomrule
        \label{tab:baseline}
        \end{tabular}
        \end{adjustbox}
\end{minipage}
\hfill
\begin{minipage}[t]{0.42\linewidth}
    \centering
        \captionof{table}{Ablation study}
        \begin{adjustbox}{width=\linewidth}
        \begin{tabular}[t]{@{\hskip 1mm}lccc@{\hskip 1mm}cccccccccccc}
        \toprule
        \multirow{3}{*}{Method} 
         & \multirow{2}{*}{\shortstack[c]{MPJPE \\ \lbrack mm\rbrack}} & \multirow{2}{*}{\shortstack[c]{PA-MPJPE \\ \lbrack mm\rbrack}} & \multirow{2}{*}{\shortstack[c]{PCKh \\ @0.5}}\\
        \\
        & ($\downarrow$) & ($\downarrow$) & ($\uparrow$)
        \\
        \midrule
        Ours w/o AME & 115.2 & 67.5 &  0.38\\
        Ours w/o TPD & 116.5& 69.0& 0.38  \\
        \rowcolor{mypink!20}
        Ours & \textbf{106.5} & \textbf{65.0} & \textbf{0.43} \\
        \bottomrule
        \label{tab:ablation}
        \end{tabular}
        \end{adjustbox}
\end{minipage}
\end{center}
\end{figure}

\begin{figure*}[t]
  \centering
   \includegraphics[width=\linewidth]{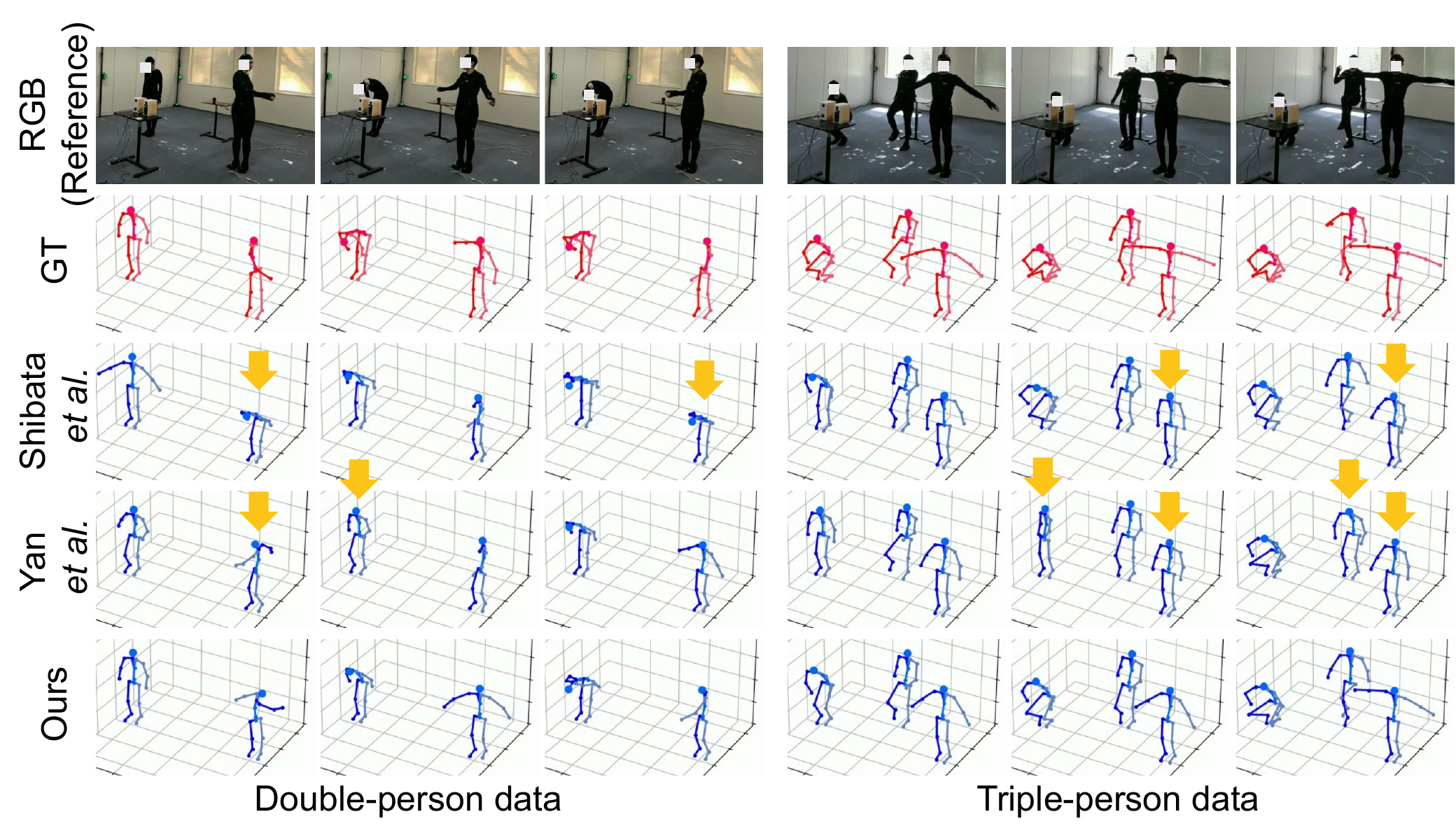}
   \caption{Qualitative results. Frames were sampled at one-second intervals for both the double-person data and the triple-person data. The yellow arrow indicates a skeleton for which pose estimation failed. }
   \label{fig:qualitative}
\end{figure*}


\subsection{Implementation Details}
\label{sec:44}

The values of 
$N_\mathrm{out}$ and $N_\mathrm{prev}$ were set to 8 and 16, respectively. 
The number of Mel filter banks $B$ for the log-Mel spectrograms was set to 128 and embedding dimension $E$ was 256.
We set the number of AME blocks and TPD blocks to $N_\mathrm{enc}=3$ and $N_\mathrm{dec}=2$, respectively.
The number of queries in the TPD is set to 135 ($M=15, N_\mathrm{out}=8$). 
For loss calculation, the weight was set to $\lambda=0.5$.
All experiments used AdamW~\cite{AdamW_loshchilov_2017} as the optimizer, with a weight decay of $1 \times 10^{-4}$.
 The number of training epochs was 500, and the learning rate was $5 \times 10^{-5}$.

\section{Experimental Results}
\label{sec:result}
We conduct seven experiments to validate the effectiveness of \modelnospace:
(1) a comparison against baseline methods;
(2) an ablation study to evaluate the effectiveness of Acoustic Multi-scale Encoder and Temporal Pose Decoder;
(3) analysis of the performance gap between single-person and multi-person settings;
(4) evaluation of the proposed attention methods in AME and TPD; 
(5) investigation of the impact of temporal window selection in multi-scale STFT;
(6) generalization to unseen environments; and
(7) a cross-modal evaluation demonstrating that \model can be successfully applied to WiFi signals with clear performance gains.

\subsection{Comparison with Baseline Methods}
\label{sec:51}

To demonstrate the effectiveness of \modelnospace, we conduct comparisons with baseline models.
We trained these models using data from two of the three groups defined in \cref{sec:41} (consisting of ten subjects), and conducted cross-subject (group) evaluation with the remaining unseen group (consisting of five subjects). The test group was rotated in a three-fold cross-validation setup, and the final evaluation results were obtained by averaging.
Table~\ref{tab:baseline} presents a quantitative comparison between \model and the baseline models. 
\model
outperforms all baselines 
across all evaluation metrics.

\cref{fig:qualitative} represents a qualitative comparison with baseline methods.
In the double-person scenario, the baseline models struggle to track dynamic ``twisting'' motions, whereas \model reconstructs them with high fidelity. 
Similarly, for the triple-person data, \model successfully estimates ``raising both arms'', a motion characterized by a small sound reflection area, even when performed alongside a walking individual. 
We hypothesize that for motions involving subtle acoustic perturbations, such as twisting or arm raising, the integration of fine-grained frequency features via the AME and the explicit modeling of inter- and intra-person dependencies in the TPD are particularly effective.

\subsection{Ablation Study}
\label{sec:52}
To demonstrate the effectiveness of our 
main technical contributions, 
 we compared two ablation settings: 
(1) excluding Acoustic Multi-scale Encoder (w/o AME), and (2) excluding Temporal Pose Decoder (w/o TPD).
In setting (1), the spectrogram $a^\mathrm{high\text{-}time}$ extracted using a single temporal window is used as the model input.
In setting (2), a total of $M$ queries, each assigned to each individual, are fed into the decoder. Each query predicts both the multi-frame pose sequence and its associated confidence scores.
In both settings, self-attention is implemented using standard self-attention.

\cref{tab:ablation} presents a quantitative comparison by evaluating the model's performance when specific components are removed. 
The results demonstrate that our complete method, which incorporates all proposed components, achieves the highest estimation accuracy across all evaluation metrics.
In addition, the results indicate that TPD contributes most significantly to the performance improvement, highlighting the importance of assigning a separate query to each predicted pose frame and extracting temporal information through cross-attention.

\subsection{Comparison by the Number of Subjects}

To analyze the performance gap between single-person and multi-person pose estimation, we compare the easiest setting (single-person) and the most challenging setting (three-person). 
Table~\ref{tab:subject} shows that
\model maintains competitive accuracy even in the multi-person setting, outperforming the baseline without significant degradation in performance.

\begin{figure}[t]
\begin{center}    
\begin{minipage}[t]{0.59\linewidth}
    \centering
        \captionof{table}{Comparison on single and triple person}
        \begin{adjustbox}{width=\linewidth}
        \begin{tabular}[t]{@{\hskip 1mm}lccc@{\hskip 1mm}cccccccccccc}
        \toprule
        & \multicolumn{3}{c}{Single-person} & & \multicolumn{3}{c}{
        Triple-person}\\
        \cmidrule{2-4} \cmidrule{6-8}
        \multirow{3}{*}{Method} 
         & \multirow{2}{*}{\shortstack[c]{MPJPE \\ \lbrack mm\rbrack}} & \multirow{2}{*}{\shortstack[c]{PA-MPJPE \\ \lbrack mm\rbrack}} & \multirow{2}{*}{\shortstack[c]{PCKh \\ @0.5}}
         & & \multirow{2}{*}{\shortstack[c]{MPJPE \\ \lbrack mm\rbrack}} & \multirow{2}{*}{\shortstack[c]{PA-MPJPE \\ \lbrack mm\rbrack}} & \multirow{2}{*}{\shortstack[c]{PCKh \\ @0.5}}\\
        \\
        & ($\downarrow$) & ($\downarrow$) & ($\uparrow$) & & ($\downarrow$) & ($\downarrow$) & ($\uparrow$) 
        \\
        \midrule
        Shibata \textit{et al.} ~\cite{listening_shibata_2023} &111.3 & 65.8 & 0.39 & & 124.5 & 73.3 & 0.39 \\
        Yan \textit{et al.} ~\cite{PETR-wifi_yan_2024} & 108.7& 64.1& 0.40&& 122.4 & 71.2 & 0.38 \\
        \rowcolor{mypink!20}
        Ours & \textbf{95.0} & \textbf{58.9} &\textbf{ 0.47}& & \textbf{111.2}&\textbf{68.1} & \textbf{0.44}\\
        \bottomrule
        \label{tab:subject}
        \end{tabular}
        \end{adjustbox}
\end{minipage}
\hfill
\begin{minipage}[t]{0.38\linewidth}
    \centering
        \captionof{table}{Comparison on self-attention methods}
        \begin{adjustbox}{width=\linewidth}
        \begin{tabular}[t]{@{\hskip 1mm}llcc@{\hskip 1mm}cccccccccccc}
        \toprule
        \multirow{3}{*}{Encoder} & \multirow{3}{*}{Decoder}
         & \multirow{2}{*}{\shortstack[c]{MPJPE \\ \lbrack mm\rbrack}} & \multirow{2}{*}{\shortstack[c]{PA-MPJPE \\ \lbrack mm\rbrack}} & \multirow{2}{*}{\shortstack[c]{PCKh \\ @0.5}}\\
        \\
        & & ($\downarrow$) & ($\downarrow$) & ($\uparrow$)
        \\
        \midrule
        Standard & MSA+ISA & 111.7 & 65.1&  0.40 \\
        TSA+FSA& Standard& 114.4 & 67.9 & 0.39 \\
        \rowcolor{mypink!20}
        TSA+FSA&  MSA+ISA  & \textbf{106.5} & \textbf{65.0} & \textbf{0.43} \\
        \bottomrule
        \label{tab:attention}
        \end{tabular}
        \end{adjustbox}
\end{minipage}
\end{center}
\end{figure}

\subsection{Effect of self-attention}
To investigate the effectiveness of the proposed Temporal Self-Attention (TSA) and Frequency Self-Attention (FSA) in the encoder, as well as Motion Self-Attention (MSA) and Interaction Self-Attention (ISA) in the decoder, we conduct ablation studies by replacing either the encoder or the decoder attention modules with standard self-attention computed over all queries.
Table~\ref{tab:attention} presents the experimental results, showing that in both the encoder and decoder settings, the models incorporating our proposed attention mechanisms consistently achieve higher performance than those using standard self-attention.

\subsection{Effect of Temporal Window Selection}
\begin{figure}[t]
\begin{center}    
\begin{minipage}[t]{0.47\linewidth}
    \centering
        \captionof{table}{Comparison on STFT window selection}
        \begin{adjustbox}{width=\linewidth}
        \begin{tabular}[t]{@{\hskip 1mm}lccc@{\hskip 1mm}cccccccccccc}
        \toprule
        \multirow{3}{*}{Window} 
         & \multirow{2}{*}{\shortstack[c]{MPJPE \\ \lbrack mm\rbrack}} & \multirow{2}{*}{\shortstack[c]{PA-MPJPE \\ \lbrack mm\rbrack}} & \multirow{2}{*}{\shortstack[c]{PCKh \\ @0.5}} \\
        \\
        & ($\downarrow$) & ($\downarrow$) & ($\uparrow$)
        \\
        \midrule

        $(L/4), (L/2), L$& 118.5 & 69.6 & 0.37 \\
        $(L/2), L, 2L$& 117.0 & 68.9 & 0.37  \\
        $L, 2L$ & 114.5 & 67.9 & 0.38\\
        \rowcolor{mypink!20}
        $L, 2L, 4L$& \textbf{106.5} & \textbf{65.0} & \textbf{0.43} \\
        \bottomrule
        \label{tab:window}
        \end{tabular}
        \end{adjustbox}
\end{minipage}
\hfill
\begin{minipage}[t]{0.5\linewidth}
\vspace{1mm}

    \includegraphics[width=\linewidth]{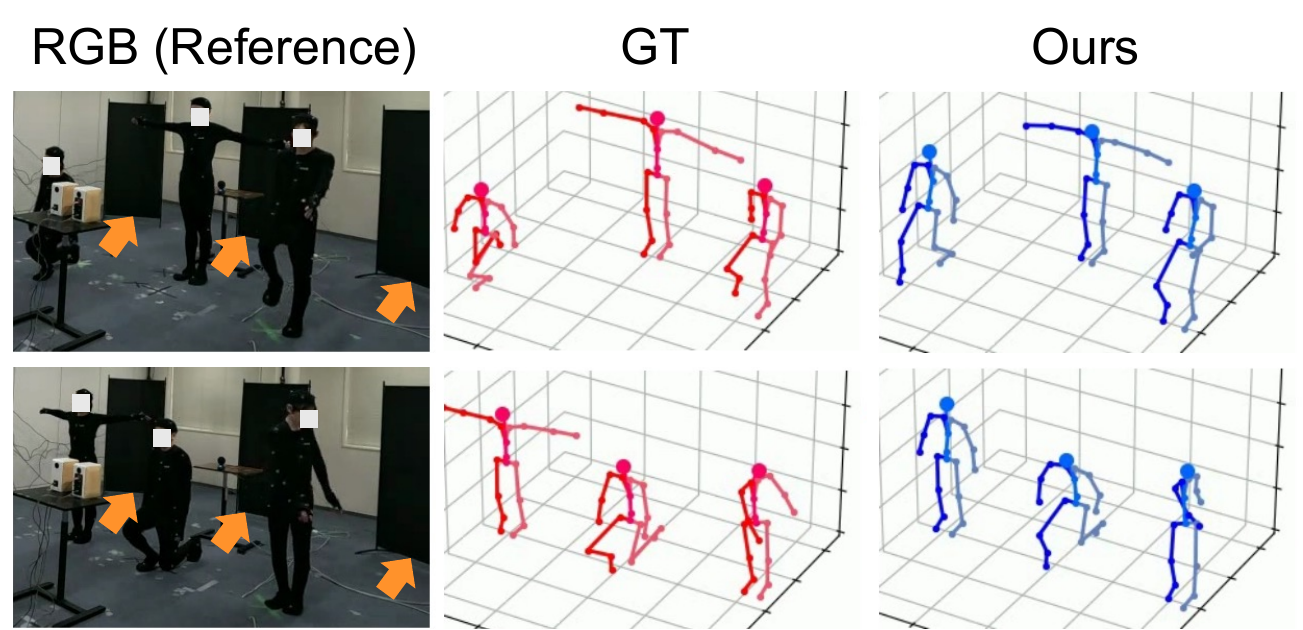}
    \captionof{figure}{Experiments in an unseen acoustically reflective environment created by black partitions (orange arrows).}
    \label{fig:wall}

\end{minipage}
\end{center}
\end{figure}

By employing window sizes of $L, 2L$, and $4L$, our AME extracts multi-resolution acoustic features. This configuration captures high-fidelity temporal dynamics aligned with the target pose frame rate ($L$) while simultaneously leveraging extended windows to extract finer frequency information.
To validate the rationale behind our window size selection, we compare our proposed $(L, 2L, 4L)$ configuration against three alternative variants:
\textbf{Shifted Resolution Range} $(L/2, L, 2L)$: This setting balances the scales around the pose-aligned resolution $L$ by including both narrower and wider windows. 
\textbf{High Temporal Resolution Focus} $(L/4, L/2, L)$: This configuration prioritizes fine-grained temporal cues at the expense of spectral depth, using only windows equal to or narrower than $L$.
\textbf{Reduced Multi-scale Variety} $(L, 2L)$: This variant evaluates the impact of limited scale diversity by employing only two window sizes.
As shown in Table~\ref{tab:window}, our $(L, 2L, 4L)$ configuration achieves the highest accuracy, confirming that this specific combination of temporal and frequency resolutions is most effective.
Furthermore, we observe that estimation accuracy decreases when using finer temporal windows at the expense of frequency resolution. This suggests that
high frequency resolution is critical to resolving the subtle acoustic variations induced by the fine-grained movements of multiple individuals.

\subsection{Generalization to unseen environments}
Generalization to environments with unseen reflection characteristics remains a challenge for sound-based pose estimation~\cite{listening_shibata_2023,sound2pose-pos_Oumi_2024}.
To evaluate the performance of \model in unseen environments, we placed multiple partitions in the room to create different reflection properties. 
As shown in~\cref{fig:wall}, \model can still estimate coarse poses even under these different reflection conditions, indicating adaptability to unseen environments.

\subsection{Cross-Modal Applicability}
\begin{figure}[t]
\begin{center}    
\begin{minipage}[t]{0.4\linewidth}
    \centering
        \captionof{table}{Quantitative comparison on the Person-in-WiFi 3D (PiW) dataset. }
        \begin{adjustbox}{width=\linewidth}
        \begin{tabular}[t]{@{\hskip 1mm}lccc@{\hskip 1mm}cccccccccccc}
        \toprule
        \multirow{3}{*}{Method} 
         & \multirow{2}{*}{\shortstack[c]{MPJPE \\ \lbrack mm\rbrack}} & \multirow{2}{*}{\shortstack[c]{PA-MPJPE \\ \lbrack mm\rbrack}} & \multirow{2}{*}{\shortstack[c]{PCKh \\ @0.5}} \\
        \\
        & ($\downarrow$) & ($\downarrow$) & ($\uparrow$)
        \\
        \midrule
        PiW (Single) & 127.4 & 71.5 & 0.13 \\
        PiW (Multi) & 161.0  & 82.7 & 0.06  \\
        \rowcolor{mypink!20}
        Ours & \textbf{122.6} & \textbf{69.1} & \textbf{0.31} \\
        \bottomrule
        \label{tab:piw3d}
        \end{tabular}
        \end{adjustbox}
\end{minipage}
\hfill
\begin{minipage}[t]{0.55\linewidth}
\vspace{1mm}
    \includegraphics[width=\linewidth]{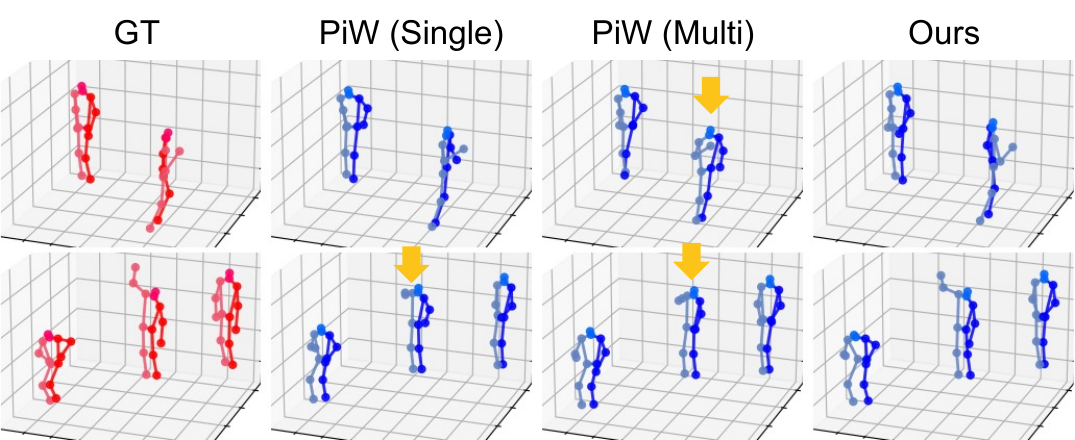}    \captionof{figure}{Visualization of predictions on PiW dataset. 
  Yellow arrows indicate failure cases. }
    \label{fig:piw_vis}
\end{minipage}
\end{center}
\end{figure}

To evaluate the generality of \model beyond acoustic sensing, we apply its modality-agnostic components to Person-in-WiFi 3D (PiW)~\cite{PETR-wifi_yan_2024}, a WiFi-based multi-person 3D pose estimation benchmark. We compare the original single-frame PiW model, denoted as PiW (Single), with PiW (Multi), a diagnostic variant extended to input and predict a 0.27-s sequence (T=4), and our method, which also uses T=4. Because WiFi channel state information (CSI) is not a waveform, we omit the waveform-specific Multi-Scale STFT and apply AME self-attention and the full TPD directly to the CSI features. All methods are trained and evaluated on the PiW dataset under the same protocol. As shown in Table~\ref{tab:piw3d}, our method outperforms both PiW (Single) and PiW (Multi).
The performance degradation of PiW (Multi) indicates that simply extending a single-frame model to sequence prediction is insufficient for modeling frame-specific signal-to-pose correspondences and temporal dependencies. In contrast, the proposed self-attention mechanism and TPD-based temporal modeling remain effective in the WiFi domain, demonstrating the cross-modal applicability of our architecture.
The qualitative results in Fig.~\ref{fig:piw_vis} further show that our method estimates dynamic poses, such as walking and hand waving, more accurately than both PiW (Single) and PiW (Multi).

\section{Conclusion}
\label{sec:conclusion}
In this paper, we presented the first approach for multi-person pose estimation using active acoustic sensing with a pair of speakers and a microphone.
To address the complexity of acoustic signals arising from multiple people, we introduced the Acoustic Multi-scale Encoder, which leverages spectrograms with diverse temporal and frequency characteristics. Furthermore, we employ the Temporal Pose Decoder to capture the relationship between a pose in each frame and the corresponding acoustic signals. 
These contributions 
improved the robustness of \modelnospace, allowing it to outperform baselines across all evaluation metrics.

Because sound-based human pose estimation is still at a very early stage of research, one of the general challenges in this task is its real-world deployment. We consider that one of the reasons for this is that large-scale, comprehensive datasets have not yet been constructed. We believe that our work in constructing an original dataset and developing methods for multi-person pose estimation will contribute to pioneering this field.
For future research, we plan to further advance sound-based pose estimation by evaluating our \model across multiple experimental environments and improving the robustness and generalization capability of the model.

\section*{Acknowledgments.}
This work was partially supported by
JSPS Grant-in-Aid for Challenging Research (Exploratory)
24K22296, JST FOREST Program JPMJFR242I, and 
JST BOOST JPMJBS2409.

%
%
\bibliographystyle{splncs04}
\bibliography{main}
\end{document}